\pdfoutput=1
\documentclass[11pt]{article}
\usepackage[table]{xcolor}
\usepackage[preprint]{acl}

\usepackage{times}
\usepackage{latexsym}
\usepackage{booktabs}
\usepackage{multirow}
\usepackage{soul} 

\usepackage[T1]{fontenc}

\usepackage[utf8]{inputenc}

\usepackage{microtype}

\usepackage{inconsolata}

\usepackage{graphicx}
\usepackage{colortbl}
\usepackage[normalem]{ulem}
\useunder{\uline}{\ul}{}
\title{Generative Debunking of Climate Misinformation}

\author{Francisco Zanartu\quad Yulia Otmakhova\quad John Cook\quad Lea Frermann \\
  The University of Melbourne, Australia \\
  \texttt{\{francisco.zanartu, y.otmakhova, jocook, lea.frermann\}@unimelb.edu.au}}

\begin{document}
\maketitle
\begin{abstract}
Misinformation about climate change causes numerous negative impacts, necessitating corrective responses. Psychological research has offered various strategies for reducing the influence of climate misinformation, such as the fact-myth-fallacy-fact-structure. However, practically implementing corrective interventions at scale represents a challenge. Automatic detection and correction of misinformation offers a solution to the misinformation problem. 
This study documents the development of large language models that accept as input a climate myth and produce a debunking that adheres to the fact-myth-fallacy-fact (``truth sandwich'') structure, by incorporating contrarian claim classification and fallacy detection into an LLM prompting framework. We combine open (Mixtral, Palm2) and proprietary (\mbox{GPT-4}) LLMs with prompting strategies of varying complexity. Experiments reveal promising performance of GPT-4 and Mixtral if combined with structured prompts. We identify specific challenges of debunking generation and human evaluation, and map out avenues for future work. We release a dataset of high-quality truth-sandwich debunkings, source code and a demo of the debunking system.\footnote{{\bf Dataset}: \url{https://huggingface.co/datasets/fzanartu/CARDSexamples}, {\bf Code}: \url{https://huggingface.co/spaces/fzanartu/flicc-agent/tree/main}, {\bf Interactive demo}: \url{https://huggingface.co/spaces/fzanartu/flicc-agent}}

\end{abstract}

\section{Introduction}


Misinformation about climate change causes numerous negative impacts such as reducing climate literacy \citep{taddicken2023climate}, polarizing the public \citep{cook2017neutralizing}, and cancelling out attempts to communicate factual information \citep{van2017inoculating}. Psychological research has offered various strategies for reducing the influence of climate misinformation, including the recommended approach of adhering to the fact-myth-fallacy-fact structure of a debunking (Figure~\ref{fig:example}; \citet{lewandowsky2020debunking}). 

\begin{figure}
    \centering
    \begin{small}
    \begin{tabular}{p{0.46\textwidth}}\toprule
        \rowcolor{darkgray}
        \textcolor{white}{As far as green plants are concerned, CO2 is not a pollutant, but part of their daily breadlike water, sunlight, nitrogen, and other essential elements.}
        \\\midrule
        {\bf Fact} While increased CO2 levels can enhance plant growth, they also have negative impacts on ecosystems, such as promoting the spread of invasive species and increasing the severity of plant diseases. Moreover, elevated CO2 levels contribute to climate change, leading to record high temperatures, ocean heat, and sea level rise.\\
        {\bf Myth}  Green plants require CO2 for growth, similar to how humans need water and food. Plants thrive at higher CO2 levels, which existed in the past and are reproduced in some greenhouses, resulting in improved growth and yields.\\
        {\bf Fallacy} This argument oversimplifies the complex relationship between CO2 and plant growth, ignoring the negative impacts of increased CO2 levels on ecosystems and the overall climate system. While it's true that CO2 is essential for plant growth, artificially elevating its levels in greenhouses does not account for the broader consequences of climate change, such as heat stress, drought, and extreme weather events, which have negative impacts on agriculture and ecosystems.\\
        {\bf Fact} Although higher CO2 levels boost plant growth, they also facilitate invasive species and disease spread, and exacerbate climate change effects, including temperature, ocean heat, and sea level rise.\\\bottomrule
    \end{tabular}
    \end{small}
    \caption{An example input myth (top, dark gray) and fact-myth-fallacy-fact (``truth sandwich'') debunking generated by our model (bottom).}
    \label{fig:example}
\end{figure}

However, while psychological research provides best-practices for debunking, practically neutralising misinformation in real-world conditions is challenging. False information spreads faster and deeper than factual information on social media, making it difficult to counter misinformation before it has already done damage \citep{vosoughi2018spread}. In order to be effective, corrective interventions  need to be deployed at scale and faster than misinformation can spread. Automatic detection and correction of misinformation, a goal described as the “holy grail of fact-checking” \citep{hassan2015quest}, offers a solution to this challenge. 

This paper presents efforts towards the completion of this “holy grail” by synthesising generative AI with past research on climate contrarian claim classification and fallacy detection, in an approach we call {\it generative debunking}. This approach adopts elements of the 4D framework \citep{cook20244d} which involves detecting, deconstructing, debunking, and deploying corrective interventions. Specifically, we build upon the CARDS (Computer Assisted Recognition of Denial \& Skepticism) classifier which was developed to detect specific contrarian claims about climate change \citep{coan2021computer,rojasaugmented}, and the FLICC model \citep{zanartu2024detecting} that detects fallacies in climate misinformation, such as Fake experts, Logical fallacies, Impossible expectations, Cherry picking, and Conspiracy theories \citep{cook2020deconstructing}. 

Specifically, we implement our {\it generative debunking} framework by testing the ability of three unique combinations of prompting strategies of varying complexity with large language models (LLMs) of different size (Section~\ref{sec:models}) to produce a structured and psychologically grounded ``truth sandwich'' debunking for a myth~(Figure~\ref{fig:example}, Section~\ref{sec:background}). We evaluate the quality of the produced debunking~(Sections~\ref{sec:evaluation}, \ref{sec:results}), identifying a lack of factuality and relevancy as a critical shortcoming even with the latest LLMs. In Section~\ref{sec:discussion} we discuss challenges of generating valid debunkings and their evaluation, and opportunities for future research.

\section{Background}
\label{sec:background}
\paragraph{Psychologically effective debunking}
Psychological research recommends that debunkings should adopt the fact-myth-fallacy-fact structure \citep{lewandowsky2020debunking}. The fact should have the same explanatory relevance as the misinformation \citep{ecker2010explicit, seifert2002continued}. For example, if the myth was ``the sun is causing global warming'', the fact should specify the actual cause (e.g., ``CO2 emissions are causing global warming'').

On the question of whether a debunking should mention the myth that is being refuted, there has been some speculation that debunkings should avoid mentioning misinformation lest the retraction causes a counterproductive ``backfire effect'' where belief in the myth is inadvertently strengthened \citep{nyhan2010corrections}. However, researchers have found difficulty in replicating the backfire effect \citep{wood2019elusive}. Rather, it is recommended that a debunking should repeat the misinformation once, as one repetition of the myth is beneficial to belief updating \citep{ecker2017reminders}. However, further repetition of the misinformation should be avoided, as it makes information appear true, a phenomenon known as the illusory truth effect \citep{fazio2015knowledge}.

Next, the logical or argumentative fallacies underlying the misinformation should be explained \citep{cook2017neutralizing}.  Explaining fallacies is powerful as they are not domain specific, empowering recipients to see the same fallacies in other topics \citep{schmid2019effective}. Explaining fallacies has also been shown to be effective in reducing the influence of misinformation regardless of whether the correction comes before or after encountering misinformation \citep{vraga2020testing}. Incorporating fallacy explanations is especially important for more nuanced forms of misinformation such as paltering or cherry picking, which can involve truthful statements that are nonetheless misleading \citep{lewandowsky2017letting}. For fallacy detection, this research relies on the FLICC framework that summarises the five techniques of science denial—Fake experts, Logical fallacies, Impossible expectations, Cherry picking, and Conspiracy theories \citep{cook2020deconstructing}.

In a debunking, the fact should be repeated again at the end. Wrapping facts at the start and end of a correction is known as a ``truth sandwich'' \citep{konig2023debunking, sullivan2018instead}. Information that is presented first and last is usually remembered best due to the primacy and recency effects \citep{jahnke1965primacy}. The repetition also makes the fact more likely to be believed by recipients \citep{fazio2020effect}. We adopt the four-layer ``truth sandwich'' (Figure~\ref{fig:example}) to structure the output of the models described below.

\paragraph{Automatic debunking} Automatic fact checking has attracted substantial interest in NLP, however the bulk of approaches falls short of generating a free-text justification by casting the problem as veracity prediction (a classification task)~\cite{guo2022survey}. Some works explain their veracity labels either by analyzing model-internal configurations that lead to a particular prediction, or by extracting explanatory facts from supporting or refuting documents~\cite{kotonya-toni-2020-explainable}, or generating it with LLMs~\cite{hsu-etal-2023-explanation}. A separate line of work studied automatic logical fallacy detection~\cite{jin-etal-2022-logical,alhindi-etal-2022-multitask}. To the best of our knowledge we are the first to integrate fallacy detection into an end-to-end system for psychologically grounded, structured debunking.

\section{Data}
Our automated debunking system leverages various public datasets and a novel dataset specifically curated to provide gold-standard examples of the type of debunkings we are aiming for.

\paragraph{FLICC test set}
The FLICC test set consists of 256 samples across 12 logical fallacies. These samples were used to report results for the FLICC model \citep{zanartu2024detecting}, but they are not part of its training knowledge. We also randomly sample 20 instances from this data set for model evaluation (Section~\ref{sec:evaluation}).

\paragraph{CLIMATE-FEVER dataset}
The CLIMATE-FEVER dataset~\cite{diggelmann2020climatefever} encompasses 1,535 real-world climate change-related claims. Each claim is associated with five manually annotated evidence sentences from English Wikipedia, which either support the claim, refute it, or contain insufficient information for claim validation. Our study exclusively utilises the false (refuted) claims from this dataset, and we employed the CARDS classifier \citep{rojasaugmented} to automatically label these instances with their misinformation category. The resulting refutations were used as additional context in fact generation (Section~\ref{ssec:mixtral}).

\begin{figure*}[ht]
  \includegraphics[width=\textwidth]{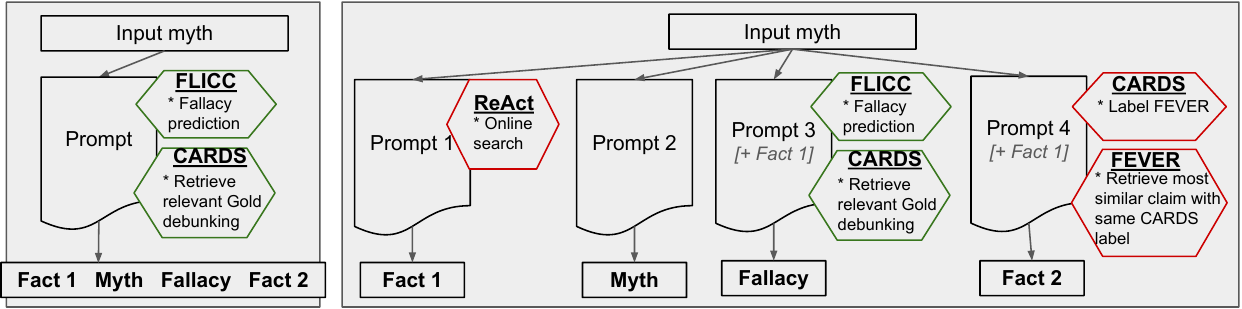}
  \caption{Overview of our dynamic prompting approaches. Left: Single prompt with dynamic fallacy prediction (FLICC) and example retrieval (CARDS). Right: Structured prompt with additional ReAct component (Fact 1) and FEVER evidence retrieval (Fact 2). External resources are shown as diamonds, and shared components between the two approaches are highlighted in green.}
  \label{fig:flowchart}
\end{figure*}

\paragraph{CARDS-examples dataset}
Additionally, we have developed a dataset of gold-standard truth-sandwich debunkings and gold fallacy labels for 62 instances of misinformation, referred to as the CARDS-examples dataset. The debunkings were created by a misinformation expert who has taught climate debunking in a Massive Open Online Course that has received over 51,000 enrolments, and is a co-author of this paper. We use this data to retrieve relevant examples for in-context learning (Sections~\ref{ssec:palm2} and~\ref{ssec:mixtral})\footnote{We include this dataset as supplementary material, and will make it public upon acceptance.}.

\section{Generative Debunking}
\label{sec:models}
We present our generative debunking approaches in order of increasing complexity, ranging from a single generic prompt (section~\ref{ssec:gpt4}), over a single prompt with myth-specific external information (section~\ref{ssec:palm2}; Figure~\ref{fig:flowchart}, left), to a structured approach that prompts individually for each layer in the debunking (section~\ref{ssec:mixtral}; Figure~\ref{fig:flowchart}, right). We apply the simpler approaches with stronger LLMs and vice versa.


The {\bf single generic prompt} is a complex end-to-end instruction for the underlying LLM that implicitly requires the model to perform fallacy detection, careful formatting, and fact retrieval. While it is easy to implement, it is limited to the LLM's internal knowledge and requires a powerful (and expensive) model to compensate for the prompt complexity and produce good-quality output. We test it with GPT4.

The {\bf single prompt with myth-specific external information} incorporates external information and guided examples into a single end-to-end prompt. It is still limited to the LLM's internal knowledge for fact retrieval. It is augmented with an external model for reasoning tasks that is not financially expensive to run. We test this more expressive prompt with a smaller, open source LLM (Palm2).

The {\bf structured prompt with myth-specific external information} involves more complex code that, in exchange, simplifies the generative debunking task into simpler subtasks. It incorporates a tool for searching the internet for specifics about certain topics and is also paired with the external reasoning model mentioned above. This most expressive prompt is combined with Mixtral.

Given the trade-off between LLM complexity and prompt complexity, we anticipate similar results among all three approaches. A more systematic experiment that disentangles the effects of LLM choice and prompting strategy is left for future work, noting that the general tendencies reported here are likely to persist, while the specifics of direct LLM comparisons tend to be short-lived with the rapid development of the technology.

\subsection{Single prompt, no context (GPT4)}
\label{ssec:gpt4}
We construct a single, comprehensive prompt which assigns the LLM the role of a climate change analyst as an expert persona~\cite{salewski2023incontext}. The instructions explain each layer in the sandwich debunking, as well as the FLICC taxonomy of logical fallacies, requesting the Fallacy component of the debunking to refer to one of the options in the taxonomy. The prompt concludes with a static example, irrespective of the given input text. Table~\ref{app:tab:gpt4} in the appendix lists the full prompt. We use this prompt with GPT-4-turbo-preview (gpt-4-0125-preview)~\cite{openai2024gpt4}, the most competitive LLM available at the time of writing. Preliminary experiments showed that weaker open-source LLMs like Palm2 when presented with this prompt, produce debunkings that are incoherent in content and/or do not comply with the truth sandwich structure.

\subsection{Single prompt, with context (Palm2)}
\label{ssec:palm2}

Hypothesizing that LLMs benefit from myth-specific context and examples, we built on the prompt presented in section~\ref{ssec:gpt4} to add dynamic context relevant to the input myth (see Figure~\ref{fig:flowchart} (left) for an illustration). First, rather than including the full FLICC taxonomy in the prompt, we now call the FLICC model~\cite{zanartu2024detecting} and dynamically insert its fallacy prediction along with the definition of the predicted fallacy, which explains how the myth misleads. Secondly, we incorporated a dynamic example into the prompt. Secondly, rather than relying on a fixed example, we now retrieve specialised instructive examples of myths with human labelled fallacies and their associated gold-standard debunking. Specifically, we encode the input myth and all myths in the CARDS-examples dataset tagged with the same type of logical fallacy using sentence-transformers\footnote{Sentence-transformers model \url{https://huggingface.co/sentence-transformers/all-MiniLM-L6-v2}}~\cite{reimers-2019-sentence-bert}. Subsequently, we select the example with the highest cosine similarity to the input myth. This selected example is then integrated into our prompt, which is now tailored to the input myth. Table~\ref{app:tab:palm2} in the appendix lists the full prompt. We use this prompt with Palm2 (text-bison-001)~\cite{anil2023palm}.

\subsection{Structured prompt, with context (Mixtral)}
\label{ssec:mixtral}

Finally, we experiment with an approach that splits the single end-to-end prompt in \ref{ssec:palm2} into four separate prompts, one per component of the output debunking as illustrated in Figure~\ref{fig:flowchart} (right). We used these prompts with Mixtral-8x7B-Instruct-vO.1~\cite{jiang2024mixtral}.

\paragraph{Layer 1: FACT.}

To encourage a specific response (rather than information that's broadly relevant to the myth), we employ a ReAct-style agent~\cite{yao2022react} equipped with an internet search tool. This agent prompts the model to reframe the misinformation as a climate change-related query, stimulating factual investigation, and utilises this query to retrieve additional information in order to enrich its response with specific facts.

We use the ReAct implementation from the LangChain library\footnote{ReAct prompt \url{https://python.langchain.com/docs/modules/agents/agent_types/react/}} with the DuckDuckGo\footnote{Web search tool \url{https://python.langchain.com/docs/integrations/tools/ddg/}} search engine and default parameters, which extracts textual content from the top five query results. From there, we rely on the LLM capabilities to distil this text and summarize the most factual information within two sentences or less than 30 words, following the guidelines of the ReAct prompt~(Table~\ref{app:tab:ReAct} in the Appendix). The resulting text is shared within prompts for layer 3 and layer 4 (see Figure~\ref{fig:flowchart}) to preserve coherence and consistency across all involved layers. The text is transmitted as dynamic content in prompt 3 (see Table~\ref{app:tab:fallacy}) and prompt 4 (see Table~\ref{app:tab:factA} and~\ref{app:tab:factB}).

\paragraph{Layer 2: MYTH.}
We directly prompt the LLM to paraphrase and summarise the input myth within 30 words in clear and concise language without adding additional information. The full prompt is in Table~\ref{app:tab:summ} in the Appendix. 

\paragraph{Layer 3: FALLACY.}
We adapt the strategy presented in section \ref{ssec:palm2}, taking inspiration from chain-of-thought methodology. We retrieved the logical fallacy prediction from the FLICC model, alongside the fallacy definitions and corresponding gold-standard debunkings employing the same process described in section \ref{ssec:palm2}. We added special delimiters to separate system messages from instructions. We also integrated factual information obtained from the initial fact layer (\textbf{Layer 1: FACT}), providing general context to facilitate coherence between this now two separated prompts. The refined prompt instructs the LLM to generate two sentences that identify the fallacy contained in the input myth and elucidate its inaccuracies by connecting it to factual evidence, showing how it distorts the reality (full prompt in Table~\ref{app:tab:fallacy} in the Appendix).

\paragraph{Layer 4: FACT.}
The final module reinforces the fact of layer one, with the opportunity to introduce supplementary information for enhanced comprehension. The process here is as follows:

\begin{enumerate}
    \item Predict the label of the input myth with the CARDS classifier.
    \item Identify all CLIMATE-FEVER claims that share the same label as our input myth.
    \item Identify the claim that is most similar to the input myth as in layer 3 by computing cosine similarity in sentence embedding space. We finally obtain the five manually annotated evidence sentences with highest cosine similarity that explain why the claim is refuted.
    \item Add these five sentences into our prompt as potential sources of new communication elements, while relying on the LLM's capabilities to determine their relevance.
\end{enumerate}

\section{Evaluation}
\label{sec:evaluation}

\begin{table*}[th]
\setlength{\tabcolsep}{4pt}
\begin{small}
\begin{tabular}{ll|ccc|ccc|ccc}
\toprule
&&\multicolumn{3}{c|}{FACT1}&\multicolumn{3}{c|}{FACT2} & \multicolumn{3}{c}{FALLACY}\\
  & & Agreement  &   Cohen's $\kappa$ & Gwet & Agreement  &   Cohen's $\kappa$ & Gwet & Agreement  &   Cohen's $\kappa$ & Gwet \\\midrule
\multirow{2}{*}{GPT4} & With expert & 43\% & 0.17 & 0.16 & 60\% & 0.32 & \textit{0.43} & 75\% & 0.55 & \textbf{0.66}\\ 
& Non-experts & 38\%	& 0.04	& 0.1	& 47\%	& 0.08	& 0.25	& 68\%	& 0.36	& \textit{0.58}\\\midrule

\multirow{2}{*}{Palm2} & With expert & 43\%	& 0.12	& 0.17	& 45\%	& 0.16	& 0.2	& 50\%	& 0.2  &	0.28\\ 
& Non-experts & 42\%	& 0.14	& 0.15	& 43\%	& 0.2	& 0.16	& 53\%	& 0.29	& 0.31\\\midrule

\multirow{2}{*}{Mixtral} & With expert & 38\%	&0.17	& 0.09	& 41\%	& 0.2	& 0.14	& 67\%	& 0.37	& \textit{0.55}\\ 
& Non-experts & 55\%	& 0.26	& 0.36	& 58\%	 & 0.29	& \textit{0.42}	 & 61\%	 & 0.17	 & \textit{0.5}
\\\bottomrule
\end{tabular}
\end{small}
\caption{Inter-annotator agreement on the scores assigned to facts (columns FACT1 and FACT2) and fallacies (FALLACY), as measured by accuracy-type agreement, Cohen's $\kappa$, and Gwet's AC1 (``Gwet'' in the table). Cases with moderate agreement (Gwet's AC1 > 0.4) are in \textit{italics}; substantial agreement (Gwet's AC1 > 0.6) are in \textbf{bold}.}
\label{tab:iaa}
\end{table*}

We evaluate the three models on their debunking of 20 climate myths, which were taken from the test data set of the FLICC model (i.e., instances that none of our models has ever been exposed to during training).

We devise a structured validation approach where Fact 1, Fallacy, and Fact 2 of the debunking are separately rated on a scale of 1 (major flaws), 2 (minor flaws) and 3 (excellent).\footnote{We also checked if outputs adhere to the sandwich structure, but all models complied 100\% of the time and we disregard this score going forward.} The rating criteria are based on the rubrics that were used for evaluating students' work during the Massive Open Online Course on climate debunking. The full evaluation instructions are included in Appendix (Table~\ref{app:tab:rubric}). The annotators were also provided the list of fallacies with their explanations and examples.

Four authors of this paper, one of whom is an expert in climate misinformation, independently evaluated 60 debunkings (20 myths debunked by the three models). The annotators were blind to the model which generated a particular debunking. The inter-annotator agreement is shown in Table~\ref{tab:iaa} (separately for each model). We report the averaged agreement between each pair of three non-expert annotators (Non-experts), and the averaged agreement between each of the non-expert annotators with the expert (With expert). We report the common inter-annotator agreement metrics such as percent agreement (the percentage of cases where both annotators assigned the same score) and Cohen's $\kappa$ \citep{cohen1960coefficient}. In addition, we use Gwet's AC1 \citep{gwet2001handbook}, which is a more reliable metric for data with potentially skewed distribution, as in our case where the lowest score (1) is underrepresented.

Overall, across all model outputs, we observe poor agreement for facts (both with the expert, and between the non-expert annotators), and a substantially better agreement for fallacies, which highlights the difficulty of judging the correctness and relevancy of the facts used in debunking for people who are not experts in climate misinformation. Among the models, the agreement for Palm2 outputs is the lowest, failing to reach moderate agreement even for the easier task of fallacy classification. In particular, non-expert annotators demonstrated widely different and inconsistent behaviours in judging Palm2 samples. On further examination, these samples tend to contain more generic information than those produced by other models, which probably caused some of the annotators grade them more positively as generally relevant and correct, while the others punished them for lack of specificity and direct relatedness (see the middle column in~Table~\ref{tab:examples} for an example of this). Another agreement abnormality is that while in general the agreement with the expert is higher than between the non-expert annotators, this is reversed for facts produced by the Mixtral model, where the non-expert agreement was substantially higher than for the other two models, while the agreement with expert was significantly lower than for other models. In other words, the non-expert annotators demonstrated a consistent behaviour which went against the expert's judgements. While we discuss these issues in more detail in Section~\ref{sec:discussion}, here we note that this is probably explained by the tendency of Mixtral's outputs to contain more specific (but not necessarily correct and relevant) facts, which can fool the non-expert annotators to make uniform judgements (left column in ~Table~\ref{tab:examples}). 

Finally, the averaged scores of the non-expert annotators are systematically more optimistic than the assessment by the expert alone (Table~\ref{tab:results} middle vs right), suggesting that the non-expert annotators overestimate their own expertise, and/or have a stronger tendency to believe the seemingly plausible facts generated by the models. Moreover, non-expert annotators strongly prefer the Mixtral outputs, while the expert judged their quality as lower than that of the GPT4 outputs (for facts) or on par with it. This is especially prominent for the facts generated by the Mixtral model, which were given significantly higher scores by non-expert annotators than by the expert. The reason behind that might be that the inclusion of external information in the Mixtral model led to generation of more specific, detailed facts, which sound more convincing for the non-expert annotators (an example of this is shown in Table~\ref{tab:examples}). These facts, however, even if they are correct, might not address the myth properly: as we mentioned in Section~\ref{sec:background}, the facts should have explanatory relevance, i.e. correct the misinformation by presenting information directly relevant. We find that it is difficult both for the models to generate such relevant facts, and for the non-expert annotators to distinguish irrelevant facts from topically related ones. We show some examples of such cases where the  non-expert annotators failed to detect an incorrect or irrelevant fact below in Section~\ref{sec:discussion}, where we discuss challenges of evaluating climate myth debunkings.

\begin{table*}[th]
\setlength{\tabcolsep}{4pt}
\begin{small}
\begin{tabular}{ p{5cm} | p{5cm}| p{5cm} }
\toprule
 {\bf GPT4} & {\bf Palm2} & {\bf Mixtral} \\\midrule
{\bf Fact} The Sun's output has been studied extensively and shows slight variations over an 11-year cycle. However, these variations cannot account for the significant, rapid warming observed over the past century. \textcolor{blue}{[Non-expert score: 1.33 (1, 1, 2); expert score: 2]} & {\bf Fact} Sunspots have been declining over the past 40 years, while global temperatures have risen. \textcolor{blue}{[Non-expert score: 2 (1, 3, 2); expert score: 2]} & {\bf Fact} While the sun does influence Earth's climate, it is not the primary cause of modern climate change. Over the past 200 years, only about 0.3$^{\circ}$C of the observed warming can be attributed to solar irradiance, according to NASA. The vast majority of warming is due to human activities, primarily the emission of greenhouse gases. \textcolor{blue}{[Non-expert score: 3 (3, 3, 3); expert score: 2]} \\
{\bf Myth} \_\_\_ & {\bf Myth} \_\_\_ & {\bf Myth}\_\_\_ \\
{\bf Fallacy} This argument commits the single cause fallacy by suggesting sunspots alone explain recent global warming, ignoring the overwhelming evidence of greenhouse gas emissions' impact on Earth's climate. \textcolor{blue}{[Non-expert score: 2.67 (2, 3, 3); expert score: 2]} &  {\bf Fallacy} This argument ignores the evidence that the sun has been cooling over the past few decades of global warming. It also assumes that solar activity is the only one driver for global warming, which ignores other factors such as man-made greenhouse gases. \textcolor{blue}{[Non-expert score: 1.67 (1, 2, 2); expert score: 3]} & {\bf Fallacy} This statement commits the single cause fallacy, attributing the overall rise in temperature over the past 200 years solely to sunspots. However, only about 0.3$^{\circ}$C of the observed warming can be attributed to solar irradiance, while the majority of warming is due to human activities, primarily the emission of greenhouse gases. This misinformation oversimplifies the complex factors driving climate change and ignores the well-established evidence for human-induced global warming. \textcolor{blue}{[Non-expert score: 3 (3, 3, 3); expert score: 3]}\\
{\bf Fact} While the Sun influences Earth's climate, the rapid warming since the Industrial Revolution far exceeds natural variability, pointing to human activities, especially greenhouse gas emissions, as the primary cause. \textcolor{blue}{[Non-expert score: 1.67 (1, 2, 2); expert score: 2]} & \textbf{Fact} Many scientific studies have found that human activity is causing global warming. \textcolor{blue}{[Non-expert score: 1 (1, 1, 1); expert score: 1]} & {\bf Fact} Sun's role in climate change is minimal; only 0.3$^{\circ}$C of warming in 200 years. Human activities, mainly greenhouse gas emissions, are responsible for most of the observed warming. \textcolor{blue}{[Non-expert score: 3 (3, 3, 3); expert score: 2]}\\ 

\\\bottomrule
\end{tabular}
\end{small}
\caption{Examples of debunkings generated by the models for the same myth ``Again the overall rise of the past 200 years is easily explained by sunspots, which is why a lot of people are nervous about cooling.'' The generated myths are abbreviated to save space. The facts generated by Palm2 tend to be generic, while those produced by Mixtral tend to contain very specific details such as \textit{only 0.3$^{\circ}$C}. Non-expert evaluator tend to be distracted by such specific details, giving Mixtral's outputs higher scores than for other models, even when the answer is incomplete, as in this example, or incorrect. On the other hand, some generic facts (top cell for Palm2) are evaluated by non-experts both as highly relevant and non-relevant. } 
\label{tab:examples}
\end{table*}

\begin{table*}[th]
\setlength{\tabcolsep}{4pt}
\begin{small}
\begin{tabular}{lc|cccc|cccc|cccc}
\toprule
&&\multicolumn{4}{c|}{All}&\multicolumn{4}{c|}{Non-expert} & \multicolumn{4}{c}{Expert}\\
  Prompt & LLM       & Fact 1  &   Fact 2 & Fact avg & Flc   & Fact 1 &   Fact 2 & Fact avg & Flc & Fact 1 &   Fact 2 & Fact avg & Flc \\\midrule
Single -Cxt & GPT4     & 2.14&	{\bf2.41}&	{\bf2.28} & 2.44& 2.22	& {\bf2.43}	& 2.33 & 2.47 & {\bf 1.90}&	{\bf 2.35}&	{\bf 2.13} & 2.35\\
Single +Cxt & Palm & 1.95&	1.86&	1.91 & 2.20& 1.98 &	1.92	& 1.95 & 2.15 &1.85&	1.70&	1.78 & 2.35\\
Struct +Cxt & Mixt & {\bf 2.23}&	2.26&	2.25 & {\bf 2.55} & {\bf2.40}	&  2.42	& {\bf2.41} & {\bf 2.60} &1.70&	1.80&	1.75 & {\bf 2.40}\\\bottomrule
\end{tabular}
\end{small}
\caption{Human ratings of the two facts, their average (Fact avg) and Fallacy (Flc) of the generated debunking by our three models. We report averaged ratings of all four annotators (left), and ratings of only non-experts (center, n=3) and a climate science expert (right, n=1).}
\label{tab:results}
\end{table*}

\section{Results}
\label{sec:results}

We now compare the quality of the debunkings generated by the three models, as rated by the annotators. An example of such debunkings is shown in Table~\ref{tab:examples}.

In Table~\ref{tab:results} we separately show the averaged results from all annotators (All), non-expert annotators (Non-expert), and the expert annotations (Expert). We observe a consistent advantage of a single, generic prompt (with GPT-4) and the structured, context-sensitive prompt (with Mixtral) over the single context-sensitive prompt (with Palm2). The most complex prompt with Mixtral performs best on Fallacy explanation, while the simple prompt with GPT4 tends to perform better on fact generation, particularly as judged by the expert. The fact that Mixtral performed comparably to GPT4 indicates the value of the structured prompting framework. We anticipate that using GPT-4 with structured prompting will lead to even higher scores.

Overall, the human scores for the detected fallacies and their explanations are higher than for the two facts at the beginning and end of the debunking, even more so for the expert scores. This may be surprising as fallacy detection and explanation is a hard task, even for humans. It also reinforces the well-known fact that even state-of-the-art LLMs hallucinate non-factual information, or remain overly generic in their output.

\section{Discussion}
\label{sec:discussion}

\paragraph{Evaluation challenges.} In Sections~\ref{sec:evaluation} and \ref{sec:results} we identified challenges with the evaluation of debunking quality by lay people: the low agreement between non-experts and with the expert; the tendency of lay people to score the outputs higher than the expert; and their tendency to over-rely on specific facts presented in the debunking. Contrary to our expectations, these problems mostly concern evaluating the facts, rather than fallacies. This is probably because for fallacy evaluation the annotators were supported by the well-structured FLICC taxonomy, while when assessing facts they had to rely purely on their own background knowledge and reasoning abilities. In particular, to correctly determine the quality of a fact used in a debunking, an annotator needed to first assess its {\it correctness}, and then evaluate its {\it relevance} to the specific myth being debunked, i.e. to decide if the fact effectively addresses the point made in the myth.

Thus, there are two significant challenges which non-experts face when evaluating the quality of facts in the debunking. First, they need to possess sufficient climate literacy to assess factual statements. For example, in response to the climate myth ``incorrect ice age predictions in the 1970s discredit climate science'', the debunking claim ``there were no legitimate scientific predictions of a coming ice age in the 1970s'' is factually incorrect. There were a handful of legitimate studies published in the 1970s predicting a possible ice age under certain conditions~\citep{rasool1971atmospheric}. A high level of climate literacy is required to make such a judgement, and in this particular case only the expert annotator gave the generated fact the lowest score of 1, while the non-expert annotators trusted the fact and evaluated it positively. Considering the propensity of LLMs to generate fluent and seemingly plausible, but incorrect facts, this presents a major challenge for evaluating climate misinformation debunkings.

Second, it seems to be even more of a challenge to evaluate the relevancy of the suggested fact to the myth. For example, a Mixtral-generated debunking in response to a myth which raises doubt about the reliability of climate models (``climate change is affected by innumerable interacting variables, atmospheric CO2 levels being just one''), included the following fact: ``Yes, the increase in atmospheric CO2 levels is a significant factor in climate change, as it traps heat and raises global temperatures, according to NASA and NOAA.'' This statement is factually correct and topically related to the myth, so the non-expert annotators assigned it the highest score. However, the fact does not address the main point of the myth, i.e. the complexity of the issue and thus the inability of the climate models to predict the future, which is why the expert judged the fact as irrelevant and gave it the lowest score. As in the example above, the annotators may tend to consider a fact to be relevant if it is specific and convincing, overlooking the missing logical connection to the myth. On the other hand, if the fact is too generic, non-expert annotators tend to vary significantly in their judgments. For example, when judging the Palm2 output ``Climate change is the long-term trend of rising global temperatures, and it is caused by human activities such as burning fossil fuels.'' which was used to debunk the myth ``Climate change is a hoax and has been rebranded multiple times to keep the fear mongering going'', some non-experts said that the fact is irrelevant, while others considered it to be simple and to the point. 

Our results and the examples above underscore the difficulty of the evaluation task, and suggest that it requires a direct involvement of a domain expert or at least their close supervision. Moreover, they highlight the importance of high reliability of the system generating the debunkings, as a non-expert user is unlikely to detect its flaws.

\paragraph{New vs. known myths.} It is an open question how well the different models perform on climate myths with differing degrees of difficulty. A number of existing debunkings would already exist online for common climate myths and be included in training data for LLMs. More novel or unaddressed myths may prove more challenging for LLMs to debunk given the lack of relevant training data. While we did not explore this issue systematically, we notice that models (especially the stronger ones such as GPT4 and Mixtral) tend to perform well on more wide-spread myths such as ``CO2 is not a pollutant but a food for plants" or ``Climate change is a hoax created by scientists and politicians to make money and control people''. On the other hand, they sometimes generate irrelevant facts and incorrectly classify the fallacy for less common myths such as ``There is no trend in hurricane-related flooding in the U.S.''. More rigorous examination of this question, as well as identifying which models are most effective at debunking different types of climate myths is a topic for further study.

\paragraph{Generating `good facts'.} In an effective debunking, the presented fact must focus on the exact same target as the myth \cite{ecker2010explicit,seifert2002continued}. Generating such facts that are not only true, but also specific and on-topic turned out to be a majory obstacle for all tested LLMs. This presents a direct challenge for current LLMs, which have a well-known tendency to producing hallucinations or platitudinal text. While we aimed to improve specificity by drawing on the FEVER data base of myth debunkings, future work will need to improve the relevance of debunkings to the myth.

\paragraph{Model vs. prompting strategy} This paper presents an exploratory study of prompting strategies in combination with LLMs of different size and ability. As such, we cannot disentangle the effects of the chosen LLMs from the prompt strategy, and doing so is an avenue for future research. Our results show that a very strong LLM (GPT4) can produce competitive debunkings even give a simple prompts with no dynamic information. On the other hand, research on automatic debunking with open-source LLMs is very much worthwhile as a widely available system that relies on commercial APIs would incur unsustainable costs. Our results suggest that structured prompts with access to external data bases can bring an advantage to this end.

\section{Conclusion}
Climate misinformation has caused severe harms in the past, and its scale and effect are expected to increase with the rise strong generative AI, rendering manual debunking infeasible. Introducing the framework of {\it Generative Debunking}, this paper incorporates psychologically grounded debunking methodology into large language models. We developed a series of prompting strategies tested with various LLMs, and validated manually by expert and non-expert annotators. Our results point to major challenges in automatic debunking and concrete directions for future work, including an improvement of generated facts in specificity and relevance as well as the challenge of validating debunking systems with non-expert annotators. We hope that our code, data sets, and findings will initiate follow-up work to advance this promising line of work.

\section*{Limitations}
As discussed at length in the paper, none of our models generates facts that are reliably of a high quality. We release our system as a research tool to stimulate follow-up work and to collect user experiences in a controlled environment. It is not currently fit for broader deployment.

Our presented evaluation is small, in terms of samples covered and annotator pool. A more thorough evaluation is needed in future work which extends both dimensions.

As acknowledged in the paper, we do not systematically study the impact of the individual prompt design decisions; nor do we exhaustively combine all prompts with all LLMs. Follow-up work will involve more careful analysis of the most useful components, also in an effort to further improve particularly the fact generation parts of our generative debunkings.

We did not evaluate our current models' abilities to distinguish input myths from fact -- but rather assumed that all input is non-factual. While detecting misinformation is outside the scope of this study, the CARDS model offers the capacity to detect online misinformation \citep{coan2021computer, rojasaugmented}. Future applications of this model may integrate online misinformation detect via the CARDS model.

\section*{Acknowledgements}
This work was supported by the Melbourne Center of AI and Digital Ethics (Seed Funding `Digital Expertise') and the Australian Research Council Discovery Early Career Research Award (Grant No. DE230100761). 

\bibliography{custom}

\newpage

\appendix

\section*{Appendix}
\label{sec:appendix}

\section{Full prompts}
\label{app:gpt4}
Table~\ref{app:tab:gpt4} provides the single, generic prompt for GPT-4 (Section~\ref{ssec:gpt4}). The single context-sensitive prompt used with PaLM 2 is shown in Table~\ref{app:tab:palm2} (Section~\ref{ssec:palm2}), while Table~\ref{app:tab:ReAct},~\ref{app:tab:summ},~\ref{app:tab:fallacy},~\ref{app:tab:factA}, and~\ref{app:tab:factB} shows the structured, context-sensitive prompts we used in combination with Mixtral (Section~\ref{ssec:mixtral}).

\begin{table*} [h]
    \centering
    \rowcolors{2}{gray!10}{white}
    \begin{small}
    \renewcommand{\arraystretch}{1.5}
    \begin{tabular}{p{0.2\linewidth}p{0.35\linewidth}p{0.35\linewidth}}
    \toprule
    \textbf{TECHNIQUE} & \textbf{DEFINITION} & \textbf{EXAMPLE} \\
    \midrule
    Ad Hominem & Attacking a person/group instead of addressing their arguments. & “Climate science can't be trusted because climate scientists are biased.”\\
    Anecdote & Using personal experience or isolated examples instead of sound arguments or compelling evidence. & “The weather is cold today—whatever happened to global warming?”\\
    Cherry Picking & Carefully selecting data that appear to confirm one position while ignoring other data that contradicts that position. & “Global warming stopped in 1998.”\\
    Conspiracy Theory & Proposing that a secret plan exists to implement a nefarious scheme such as hiding a truth. & “The climategate emails prove that climate scientists have engaged in a conspiracy to deceive the public.”\\
    Fake Experts & Presenting an unqualified person or institution as a source of credible information. & “A retired physicist argues against the climate consensus, claiming the current weather change is just a natural occurrence.”\\
    False Choice & Presenting two options as the only possibilities, when other possibilities exist. & “CO2 lags temperature in the ice core record, proving that temperature drives CO2, not the other way around.”\\
    False Equivalence & Incorrectly claiming that two things are equivalent, despite the fact that there are notable differences between them. & “Why all the fuss about COVID when thousands die from the flu every year.”\\
    Impossible Expectations & Demanding unrealistic standards of certainty before acting on the science. & “Scientists can't even predict the weather next week. How can they predict the climate in 100 years?”\\
    Misrepresentation & Misrepresenting a situation or an opponent's position in such a way as to distort understanding. & “They changed the name from 'global warming' to 'climate change' because global warming stopped happening.”\\
    Oversimplification & Simplifying a situation in such a way as to distort understanding, leading to erroneous conclusions. & “CO2 is plant food so burning fossil fuels will be good for plants.”\\
    Single Cause & Assuming a single cause or reason when there might be multiple causes or reasons. & “Climate has changed naturally in the past so what's happening now must be natural.”\\
    Slothful Induction & Ignoring relevant evidence when coming to a conclusion. & “There is no empirical evidence that humans are causing global warming.”\\
    \bottomrule
    \end{tabular}
    \end{small}
    \caption{The FLICC taxonomy of twelve logical fallacies of climate misinformation as defined in~ \citep{zanartu2024detecting}.}
    \label{app:tab:flicc}
\end{table*}

\begin{table*}[h]
\begin{small}
    \begin{tabular}{p{15.5cm}}
         \toprule
<role>
You are an expert climate analyst tasked with providing precise and concise responses to climate change misinformation using a structured format similar to a "hamburger-style" response.
< $\backslash$role>\\
\\
<instruction>
Provide precise and concise replies to climate change misinformation using a structured "hamburger-style" FACT, MYTH, FALLACY, FACT:
The model consists of the following components: (leave out the CAPITALISED: words when responding use \#\# for heading, !\#\#\#! for endmarkers, to mark the end of a response.\\\\
FACT: A 30 words or fewer fact description. Offer clear, memorable alternatives to enhance comprehension. Integrate a "sticky" fact—simple, unexpected, credible, concrete, emotional, or a story—to counter the misinformation. For example: "Arctic sea ice dropped 40\% since the '70s, hitting record lows." Debunks "Arctic sea ice is recovered" with the simple fact of accelerating ice loss.\\
MYTH: Paraphrase the misinformation in 30 words or fewer. \\
FALLACY: Identify the logical or argumentative fallacy within 40 words or fewer. Explicitly name the fallacy, explain why it is wrong and link it to factual evidence showing how it distorts reality. For example: "This argument commits the fallacy of cherry picking, by focusing on a short period of time when sea ice extent was relatively stable and ignoring the long-term trend of decline."  Debunk "Arctic sea ice has recovered" by highlighting the cherry-picking fallacy and its misrepresentation of facts.\\
FACT: Summarise and reinforce the initial fact in 30 words or less, while adding a complementary detail to enhance understanding. 
Repeat the initial fact in 30 words or fewer."\\
\\
You should categorise the underlying fallacies according to the following table from the Debunking handbook:\\
\\
<PLACEHOLDER FOR FLICC TAXONOMY>\\\\

Your task is considered complete once all the elements of the hamburger-style response have been formulated, consider and adhere to the following example.
< $\backslash$instruction>\\
\\
<example>\\
myth: Earth's climate has changed naturally before, so current climate change is natural.\\
single cause fallacy: Assuming a single cause or reason when there might be multiple causes or reasons.\\\\
response:\\
\#\# FACT: Scientists observe human fingerprints all over our climate. Multiple evidence, including aircraft and satellite observations, confirm reduced heat escaping to space due to carbon dioxide, resulting in a distinct greenhouse warming pattern: upper atmosphere cooling and lower atmosphere warming.\\
\#\# MYTH: Earth's climate has changed naturally before, so current climate change is natural.\\
\#\# FALLACY: This argument commits the single cause fallacy, falsely assuming that because natural factors have caused climate change in the past, then they must always be the cause of climate change. \\
\#\# FACT: Just as a detective finds clues in a crime scene, scientists have found many clues in climate measurements confirming humans are causing global warming. Human-caused global warming is a measured fact. !\#\#\#!\\
< $\backslash$example>\\\\
myth: \textcolor{red}{\{text\}}\\
response:\\
\bottomrule
    \end{tabular}
    \end{small}
    \caption{Single, comprehensive prompt that generates the full debunking sandwich end-to-end, including a role specification, detailed instruction and an example. While the instructions refer to the FLICC taxonomy, it contains no information specific to the input myth. Dynamic content in red.}    \label{app:tab:gpt4}

\end{table*}

\begin{table*}
\begin{small}
    \begin{tabular}{p{15.5cm}}
\toprule

[Role]:\\
You are an expert climate analyst tasked with providing precise and concise responses to climate change misinformation using a structured format similar to a "hamburger-style" response.\\\\

[Instruction]:\\
Provide precise and concise replies to climate change misinformation using a structured "hamburger-style" FACT, MYTH, FALLACY, FACT:
The model consists of the following components: (leave out the CAPITALISED: words when responding use \#\# for heading, !\#\#\#! for endmarkers, to mark the end of a response.\\\\
FACT: A 30 words or fewer fact description. Offer clear, memorable alternatives to enhance comprehension. Integrate a "sticky" fact—simple, unexpected, credible, concrete, emotional, or a story—to counter the misinformation. For example: "Arctic sea ice dropped 40\% since the '70s, hitting record lows." Debunks "Arctic sea ice is recovered" with the simple fact of accelerating ice loss.\\
MYTH: Paraphrase the misinformation in 30 words or fewer. \\
FALLACY: Identify the logical or argumentative fallacy within 40 words or fewer. Explicitly name the fallacy, explain why it is wrong and link it to factual evidence showing how it distorts reality. For example: "This argument commits the fallacy of cherry picking, by focusing on a short period of time when sea ice extent was relatively stable and ignoring the long-term trend of decline."  Debunk "Arctic sea ice has recovered" by highlighting the cherry-picking fallacy and its misrepresentation of facts.\\
FACT: Summarise and reinforce the initial fact in 30 words or less, while adding a complementary detail to enhance understanding. \\
\\
Your task is considered complete once all the elements of the hamburger-style response have been formulated, consider and adhere to the following example:\\\\

[Example]:\\
Misinformation: \textcolor{red}{\{claim\}}\\
\textcolor{red}{\{fallacy\}} fallacy: \textcolor{red}{\{definition\}}\\
example response:\\
\textcolor{red}{\{example\}} !\#\#\#!\\
\\
Remember to be as concise as the example presented before and to follow the "hamburger-style" response format:\\
\\
Misinformation: \textcolor{red}{\{text\}}\\
\textcolor{red}{\{fallacy\}} fallacy: \textcolor{red}{\{definition\}}\\
response:    \\\bottomrule
    \end{tabular}
    \end{small}
        \caption{Single, context-sensitive prompt. The instructions are identical to the generic prompt except that the FLICC taxonomy is not provided. Instead, we 1) retrieve an example that is specific to the input myth; and 2) predict the fallacy with the FLICC model as described in Section~\ref{ssec:palm2}. Dynamic content in red.}    \label{app:tab:palm2}
\end{table*}


\begin{table*}
    \begin{small}
        \begin{tabular}{p{15.5cm}}
            \toprule
            <s>[INST] You will receive a piece of misinformation related to climate change. Your task is to translate this misinformation into a climate change-related question that challenges the misinformation and prompts for factual investigation.\\
            For example, if the misinformation is: "Climate change isn't real because it's been cold this winter."\\
            The translated question could be: "How does winter weather in one location relate to the broader scientific consensus on climate change?"\\
            Be as specific as possible, ensuring the question directly addresses climate change and encourages factual investigation. You have access to the following tools:\\
            \textcolor{red}{\{tools\}}\\
            {[\textbackslash INST]}\\
            <\textbackslash s>\\
            {[INST]}\\
            Use the following format:\\
            Question: the translated question challenging the misinformation and prompting for factual investigation\\
            Thought: you should always think about what to do\\
            Action: the action to take, should be one of \textcolor{red}{[\{tool\_names\}]}\\
            Action Input: the input to the action\\
            Observation: the result of the action\\
            ... (this Thought/Action/Action Input/Observation can repeat N times)\\
            Thought: I now know the final answer\\
            Final Answer: provide a factual response to the original misinformation, limit your answer two sentences or less than 30 words. Be specific, prefer facts that contain numbers or are backed up by recognised institutions or climate experts to ensure credibility.\\
            Begin!\\
            {[\textbackslash INST]}\\
            
            Question: \textcolor{red}{\{input\}}\\
            Thought: \textcolor{red}{\{agent\_scratchpad\}}
            \\\bottomrule
        \end{tabular}
    \end{small}
    \caption{Structured, context-sensitive prompt. Layer 1: ReAct prompt with internet search capabilities. The instructions guide the retrieval of relevant facts to counter climate change myths as detailed in section~\ref{ssec:mixtral}. Dynamic content in red.} \label{app:tab:ReAct}
\end{table*}

\begin{table*}
    \begin{small}
        \begin{tabular}{p{15.5cm}}
            \toprule
            {[INST]} You are a paraphrasing system capable of providing rephrased versions of texts in clear and concise language.\\
            Paraphrase the following text in 30 words or fewer. Only refer to the text without adding additional elements or opinions.\\
            {[\textbackslash INST]}\\
            text: \textcolor{red}{\{text\}}\\
            Summary: 
            \\\bottomrule
        \end{tabular}
    \end{small}
    \caption{Structured, context-sensitive prompt. Layer 2: summarising prompt. Instructs the LLM to succinctly paraphrase and summarise the input myth in 30 words or less, maintaining clarity and conciseness without introducing extraneous details. Dynamic content in red.} \label{app:tab:summ}
\end{table*}

\begin{table*}
    \begin{small}
        \begin{tabular}{p{15.5cm}}
            \toprule
            <s>{[INST]} <<SYS>>\\
            You are a senior climate analyst, an expert in identifying and responding to climate change misinformation.\\
            <<\textbackslash SYS>>\\
            What fallacy is contained in the following climate change misinformation?\\
            misinformation: \textcolor{red}{{misinformation}} {[\textbackslash INST]} \\
            Your text contains \textcolor{red}{\{detected\_fallacy\}} fallacy. \textcolor{red}{\{detected\_fallacy\}} fallacy is \textcolor{red}{\{fallacy\_definition\}}\\ 
            <\textbackslash s>\\
            <s>{[INST]} What is the factual evidence surrounding this climate change misinformation?{[\textbackslash INST]}\\
            \textcolor{red}{\{factual\_information\}}<\textbackslash s>\\
            <s>{[INST]} Provide a precise and concise response to this climate change misinformation.\\
            In two sentences, explicitly name the fallacy, explain why it's incorrect, and link it to factual evidence showing how it distorts reality.\\
            Consider the following example before providing your answer:\\
            Misinformation: \textcolor{red}{\{example\_myth\}}\\
            Response: \textcolor{red}{\{example\_response\}}\\
            Misinformation: \textcolor{red}{\{misinformation\}}\\
            Response: \\ 
            {[\textbackslash INST]}
            \\\bottomrule
        \end{tabular}
    \end{small}
    \caption{Structured, context-sensitive prompt. Layer 3: fallacy detection and explanation prompt. Similar to single, context-sensitive prompt \ref{app:tab:palm2}, (1) we predict the fallacy using the FLICC model and complement it with its definition, (2) retrieve an specific example to the input myth, (3) add factual information from ReAct prompt in Table~\ref{app:tab:ReAct}. Dynamic content in red} \label{app:tab:fallacy}
\end{table*}

\begin{table*}
    \begin{small}
        \begin{tabular}{p{15.5cm}}
            \toprule
            <s>{[INST]}\\
            1. Reinforce the following fact and provide complementary details, if relevant, to enhance understanding.\\
            2. The output should be simple text summarizing the information in 30 words or fewer. Replace technical and complex words with simpler synonyms and delete unimportant information.{[\textbackslash INST]}\\
            Complementary details:\\
            \textcolor{red}{\{complementary\_details\}}\\
            <\textbackslash s>\\
            \# Fact:\\
            \textcolor{red}{\{fact\}}\\
            \# Summary:\\
            \\\bottomrule
        \end{tabular}
    \end{small}
    \caption{Structured, context-sensitive prompt. Layer 4: closing fact when CLIMATE-FEVER claims are found. The instruction is to reinforce the factual information obtained from ReAct promt in Table~\ref{app:tab:ReAct} with the option to add relevant complementary details retrieved from CLIMATE-FEVER dataset. Dynamic content in red.} \label{app:tab:factA}
\end{table*}

\begin{table*}
    \begin{small}
    \begin{tabular}{p{15.5cm}}
    \toprule
    <s>{[INST]} \\
    1. Reinforce the following fact and provide complementary details, if relevant, to enhance understanding.<\textbackslash s>\\
    2. The output should be simple text summarizing the information in 30 words or fewer. {[\textbackslash INST]}\\
    <\textbackslash s>\\
    \# Fact:\\
    \textcolor{red}{\{fact\}}\\

    \# Summary:
    \\\bottomrule
    \end{tabular}
    \end{small}
    \caption{Structured, context-sensitive prompt. Layer 4: closing fact without providing additional details. The instruction is the same as prompt in Table~\ref{app:tab:factA} but without providing the complementary details to the prompt. Dynamic content in red.} \label{app:tab:factB}
\end{table*}

\begin{table*}[]
\scriptsize
\begin{tabular}{p{0.1\linewidth}p{0.35\linewidth}>{\centering\arraybackslash}p{0.05\linewidth}p{0.35\linewidth}}
\toprule
Fact              & Description     & Points & Example   \\
\midrule
\multicolumn{4}{p{\dimexpr 0.9\linewidth + 3\tabcolsep\relax}}{How well does the rebuttal provide a factual alternative to the myth in a sticky and fallacy-free manner? Does it include facts and evidence to support the points made throughout the writing? Look for accurate, evidence-based, simple, credible and concrete explanations.}                                                                                                                                                                                                                                                                                                  \\
Excellent         & Includes a relevant and "sticky" fact as an alternative to the myth that is accurate and fallacy-free. Stickiness contains one or more of the following: Simple, Unexpected, Credible, Concrete, Emotional, Stories.          & 3      & "Arctic sea ice has declined by 40\% since the 1970s. The rate of decline has accelerated in recent years, with sea ice extent reaching record lows in recent years." Debunks "Arctic sea ice is recovered" with simple fact of accelerating ice loss.                                                                                 \\
Good              & Includes a relevant but "non-sticky" fact as an alternative to the myth that is accurate and fallacy-free. Non-sticky facts do not contains any of the following: Simple, Unexpected, Credible, Concrete, Emotional, Stories. & 2      & "The Earth's climate has changed throughout history, but the current warming trend is unprecedented in both its speed and its magnitude." This fact used in response to "cold weather disproves global warming" is too generic/non-specific and doesn't directly address how global warming impacts cold weather.                      \\
Needs Improvement & Includes a fact that is inaccurate, irrelevant, or contains a fallacy.                                                                                                                                                        & 1      & "Fossil fuels are the cheapest form of energy, but they are also the dirtiest." Questionable statement as in some contexts, renewables have become cheaper than fossil fuels.                                                                                                                                                          \\
Inadequate        & The fact explanation is nonsensical or doesn't include a relevant fact.                                                                                                                                                       & 0                                                                                                                                                                                                                                                                                                                                            &  \\
\toprule
Fallacy           & Description                                                                                                                                                                                                                   & Points & Example                                                                                                                                                                                                                                                                                                                                \\
\midrule
\multicolumn{4}{p{\dimexpr 0.9\linewidth + 3\tabcolsep\relax}}{Focus on the Fallacy section of the rebuttal. Did the rebuttal identify the correct fallacy and explain how the myth commits the fallacy?}                                                                                                                                                                                                                                                                                                                                                                                                                                       \\
Excellent         & The rebuttal has identified the fallacy correctly and clearly explained why the myth is incorrect, tieing it to the fact (e.g., how the fallacy distorts the fact).                                                           & 3      & "This argument commits the fallacy of cherry picking, by focusing on a short period of time when sea ice extent was relatively stable and ignoring the long-term trend of decline." Debunks "Arctic sea ice is recovered" by both explaining the fallacy of cherry picking and tieing it in with the facts.                           \\
Good              & The rebuttal has identified the fallacy correctly but hasn't accurately or clearly explained why the myth is incorrect (e.g., hasn't explained how the fallacy distorts the fact).                                            & 2      & "This argument commits the slothful induction fallacy, which is the fallacy of assuming that because there is no definitive proof of something, it must not be true." This correctly identifies the fallacy of slothful induction but doesn't accurately explain the fallacy - the explanation is closer to impossible expectations.   \\
Needs Improvement & The rebuttal has not identified the fallacy correctly or makes an incorrect statement.                                                                                                                                        & 1      & "This argument commits the false cause fallacy, falsely assuming that because two things have happened together in the past, one must have caused the other." In debunking "CO2 lags temperature", it gets fallacy wrong (should be single cause) and hence the fallacy explanation is incorrect.                                      \\
Inadequate        & The fallacy explanation is nonsensical.                                                                                                                                                                                       & 0                                                                                                                                                                                                                                                                                                                                             \\
\toprule
Structure         & Description                                                                                                                                                                                                                   & Points & Example                                                                                                                                                                                                                                                                                                                                \\
\midrule
Yes               & The rebuttal adheres to the fact-myth-fallacy-fact structure.                                                                                                                                                                 & 1      & "This argument commits the fallacy of cherry picking, by focusing on a short period of time when sea ice extent was relatively stable and ignoring the long-term trend of decline." Debunks "Arctic sea ice is recovered" by both explaining the fallacy of cherry picking and tieing it in with the facts.                            \\
No                & The rebuttal doesn't adhere to the fact-myth-fallacy-fact structure.                                                                                                                                                          & 0      & "This argument commits the slothful induction fallacy, which is the fallacy of assuming that because there is no definitive proof of something, it must not be true." This correctly identifies the fallacy of slothful induction but doesn't accurately explain the fallacy - the explanation is closer to impossible expectations. \\ 
\bottomrule
\end{tabular}
\caption{Structured validation rubric, where for Fact and Fallacy sections, 3 points is ``Excellent'', 2 points is ``Good'', 1 point is ``Needs improvement'', and 0 points represent an inadequate answer. For the Structure, 1 point corresponds to ``The rebuttal adheres to the fact-myth-fallacy-fact structure'', and 0 points are given when it does not.}
\label{app:tab:rubric}
\end{table*}

\end{document}